\documentclass[a4paper,12pt]{article}

\usepackage[utf8x]{inputenc}
\usepackage[T1]{fontenc}
\usepackage{times}
\usepackage[colorlinks=true,linkcolor=black,citecolor=black,urlcolor=blue]{hyperref}
\usepackage{geometry}
\usepackage{titlesec}
\titleformat{\section}
{\bfseries\uppercase}{\thesection.}{1em}{}
\titleformat{\subsection}
{\bfseries}{\thesection.\thesubsection.}{1em}{}

\usepackage{graphicx} 
\usepackage{multirow} 
\usepackage{cite}
\usepackage{breakurl}
\usepackage{indentfirst}
\usepackage{amsmath, amssymb, amsfonts, bm}
\usepackage{txfonts}
\usepackage{enumitem}
\usepackage{color,xcolor}
\usepackage{colortbl}
\usepackage{threeparttable}
\usepackage{subfigure}

\titlespacing*{\subsection}{0pt}{1.5em}{0.2em}

\renewcommand\eqref[1]{Equation~\ref{#1}}

\renewcommand{\thesection}{\arabic{section}}
\renewcommand{\thesubsection}{\arabic{subsection}}

\newlength{\bibitemsep}
\newlength{\bibparskip}
\let\oldthebibliography\thebibliography
\renewcommand\thebibliography[1]{%
  \oldthebibliography{#1}%
  \setlength{\parskip}{\bibitemsep}%
  \setlength{\itemsep}{\bibparskip}%
}
 
\begin{document}

\begin{center}
	\includegraphics[width=2.65in]{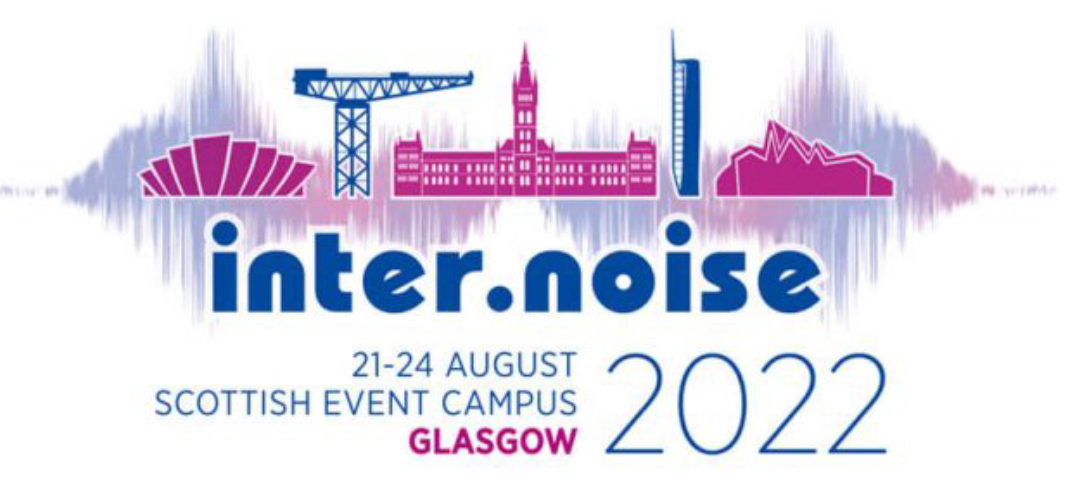}
\end{center}
\vskip.5cm

\begin{flushleft}
\fontsize{16}{20}\selectfont\bfseries
\color{black}
Performance Evaluation of Selective Fixed-filter Active Noise Control based on Different Convolutional Neural Networks
\end{flushleft}
\vskip.5cm

\renewcommand\baselinestretch{1}
\begin{flushleft}
Zhengding Luo\footnote{luoz0021@e.ntu.edu.sg}, Dongyuan Shi\footnote{dongyuan.shi@ntu.edu.sg}, Woon-Seng Gan\footnote{ewsgan@ntu.edu.sg}\\
School of Electrical and Electronic Engineering,\\
Nanyang Technological University, Singapore.\\

\vskip.5cm
Libin Zhang\footnote{zhanglibin@huawei.com}\\
Huawei Technologies Co., Ltd.\\
\vskip.5cm
Qirui Huang\footnote{huang.qirui@huawei.com}\\
Huawei International Pte. Ltd.\\

\end{flushleft}

\textbf{\centerline{ABSTRACT}}\\
\textit{Due to its rapid response time and a high degree of robustness, the selective fixed-filter active noise control (SFANC) method appears to be a viable candidate for widespread use in a variety of practical active noise control (ANC) systems. In comparison to conventional fixed-filter ANC methods, SFANC can select the pre-trained control filters for different types of noise. Deep learning technologies, thus, can be used in SFANC methods to enable a more flexible selection of the most appropriate control filters for attenuating various noises. Furthermore, with the assistance of a deep neural network, the selecting strategy can be learned automatically from noise data rather than through trial and error, which significantly simplifies and improves the practicability of ANC design. Therefore, this paper investigates the performance of SFANC based on different one-dimensional and two-dimensional convolutional neural networks. Additionally, we conducted comparative analyses of several network training strategies and discovered that fine-tuning could improve selection performance. \\
}

\section{INTRODUCTION}
\noindent
Acoustic noise problems are becoming more prevalent as the quantity of industrial equipment increases \cite{1}. The attenuation of low-frequency noises is quite difficult and expensive for passive noise control techniques such as enclosures, barriers, silencers, etc. Different from passive techniques, active noise control (ANC) involves the electro-acoustic generation of a sound field to cancel an unwanted existing sound field \cite{2}. Moreover, ANC can offer a possible lower-cost alternative for the control of low-frequency noises. Thus, it attracts much interest from the industry. When dealing with different types of noises, traditional ANC systems typically use adaptive algorithms to adjust control filter coefficients to minimize the error signal \cite{3}. Among adaptive algorithms, the filtered-X least mean square (FxLMS) and filtered-X normalized least-mean-square (FxNLMS) algorithm are commonly used since they can compensate for the delay involved by the secondary path to increase the system robustness \cite{4}.

\begin{figure}[tp]
\setlength{\abovecaptionskip}{0.cm}
\setlength{\belowcaptionskip}{-0.cm}
\centering
\centerline{\includegraphics[width=0.65\linewidth]{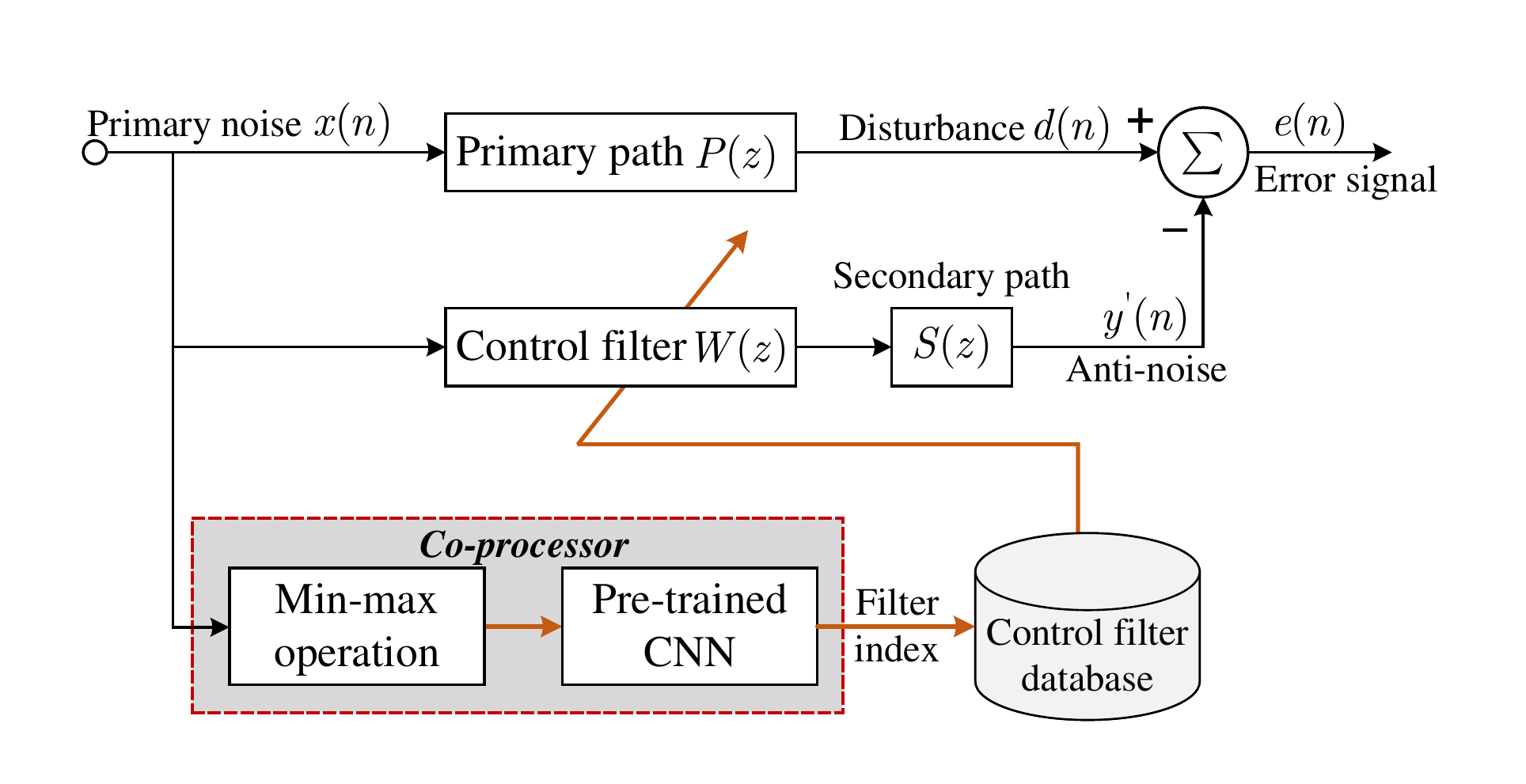}}
\caption{Block diagram of the CNN-based SFANC algorithm.}
\label{Fig 1}
\end{figure}

However, due to the least mean square (LMS) based algorithms' inherent slow convergence and poor tracking ability \cite{5}, FxLMS and FxNLMS are less capable of dealing with rapidly varying or non-stationary noises. Their slow responses to noises may impact customers' perceptions of the noise reduction effect \cite{6}. Fixed-filter ANC methods \cite{7} can be adopted to tackle slow convergence, where the control filter coefficients are pre-trained rather than adaptive updated. However, the pre-trained control filter is only suitable for a specific noise type, resulting in the degradation of noise reduction performance for other types of noises. To rapidly select different pre-trained control filters given different noise types, a selective fixed-filter active noise control (SFANC) method based on the frequency band matching was proposed in \cite{8}.

Though the SFANC method \cite{8} selects the most suitable pre-trained control filters in response to different noise types, several critical parameters of the method can only be determined through trials and errors. Considering the limitations, deep learning techniques, particularly convolutional neural networks (CNNs) \cite{9,10,21,22}, appear to be powerful in classifying noises in SFANC methods. Automatic learning of the SFANC algorithm's critical parameters based on deep learning would broaden its applications in real-world scenarios.

With the learning ability of CNN models, the SFANC algorithm can automatically learn its parameters from noise datasets and select the best control filter given different noise types without resorting to extra-human efforts \cite{11}. Additionally, a CNN model implemented on a co-processor can decouple the computational load from the real-time noise controller. Therefore, in this paper, we compared the performance of several one-dimensional (1D) CNNs and two-dimensional (2D) CNNs in the SFANC method. Also, different network training strategies are tried to choose the best one for training the networks. Experiments show that the SFANC method based on CNN not only achieves faster responses than FxLMS and FxNLMS but also exhibits good robustness. Thus, it is expected to be used for attenuating dynamic noises such as traffic noises and urban noises, etc.

\section{CNN-based SFANC Algorithm}
\noindent
The overall architecture of the CNN-based SFANC algorithm is depicted in Figure \ref{Fig 1}. Throughout the control process, the real-time controller conducts filtering to generate anti-noise while simultaneously sending the primary noise to a co-processor (e.g., a mobile phone). Given the primary noise, the co-processor employs a pre-trained CNN to produce the index for the most appropriate control filter and delivers it to the real-time controller. The controller then adjusts the control filter coefficients based on the received filter index. Notably, if the network is a 1D CNN, its input is the raw waveform \cite{25}. However, if the network is a 2D CNN, its input is the Log Mel-spectrogram \cite{19}.

\begin{figure}[tp]
\setlength{\abovecaptionskip}{0.cm}
\setlength{\belowcaptionskip}{-0.cm}
\centering
\includegraphics[width=0.5\linewidth]{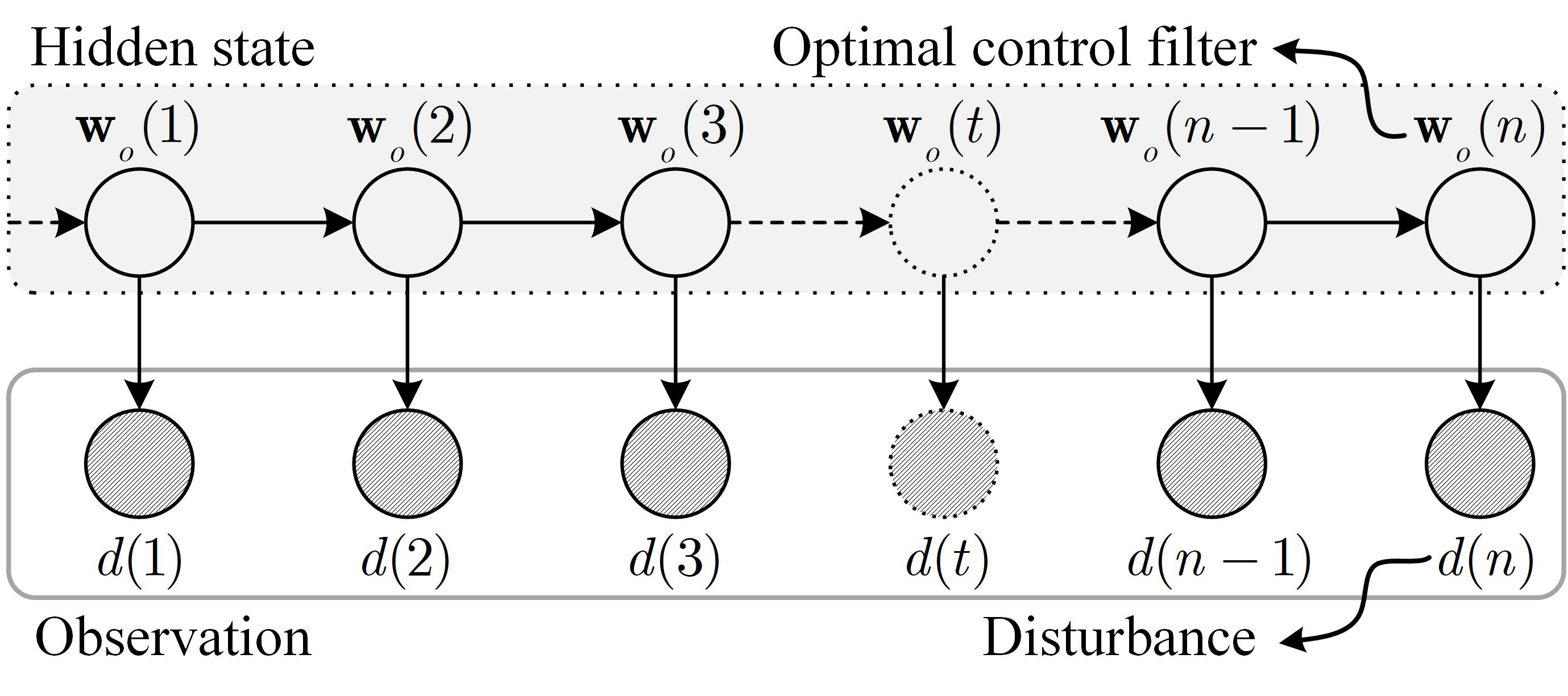}
\caption{Markov model of the ANC progress.}
\label{Fig 2}
\end{figure}

\subsection{Concise Explanation of SFANC}
\noindent
An ANC progress can be abstracted as a first-order Markov chain \cite{12} as shown in Figure \ref{Fig 2}, where $\mathbf{w}_o(n)$ represents the optimal control filter to attenuate the disturbance $d(n)$. To achieve the best noise reduction performance, the best control filter $\mathbf{w}_o(n)$ can be selected from a pre-trained filter set $\{\mathbf{w}_i\}^C_{i=1}$. Hence, the SFANC method can be represented as follows:
\begin{equation}\label{eq_s1}
    \mathbf{w}_o =  \underset{\mathbf{w}\in\{\mathbf{w}_i\}^C_{i=1}}{\mathrm{argmin}}~\mathbb{E}\bigg\{\left[d(n)-\mathbf{x}^\mathrm{T}(n)\mathbf{w}(n)\ast s(n)\right]^2\bigg\},
\end{equation}
where $\text{argmin}(\cdot)$ operator returns the input value for minimum output; $\ast$ , $\mathbf{x}(n)$, and $\mathbf{s}(n)$ represent the linear convolution, the reference signal, and the impulse response of the secondary path, respectively. The reference signal is assumed to be the same as the primary noise.

In practice, $d(n)$ is typically seen as the linear combination of $\mathbf{x}(n)$. Thus, \eqref{eq_s1} equals to
\begin{equation}\label{eq_s2}
    \mathbf{w}_{o} = \underset{\mathbf{w}\in\{\mathbf{w}_i\}^C_{i=1}}{\mathrm{argmax}} P\left[\mathbf{w}|d(n)\right] =  \underset{\mathbf{w}\in\{\mathbf{w}_i\}^C_{i=1}}{\mathrm{argmax}} P\left[\mathbf{w}|\mathbf{x}(n)\right],
\end{equation}
which means that the selected control filter is the one with maximum posterior probability in the presence of reference signal $\mathbf{x}(n)$. Moreover, according to Bayes' theorem \cite{13}, the posterior probability can be replaced with a conditional probability as \begin{equation}
    \mathbf{w}_{o} =  \underset{\mathbf{w}\in\{\mathbf{w}_i\}^C_{i=1}}{\mathrm{argmax}} P\left[\mathbf{x}(n)|\mathbf{w}\right],
\end{equation}
which predicts the most suitable control filter straight from the
primary noise $\mathbf{x}(n)$.

A classifier model $\hat{P}\left[\mathbf{x}(n)|\mathbf{w},\Theta\right]$ can be developed to approximate ${P}\left[\mathbf{x}(n)|\mathbf{w}\right]$ from the pre-recorded sampling set $\{\mathbf{x}^{j}(n), \mathbf{w}^{j}\}^{N}_j$. The $\Theta$ denotes the parameters of the classifier and can be obtained through maximum likelihood estimation (MLE) \cite{14} as
\begin{equation}
    \Theta = \underset{\Theta}{\mathrm{argmax}}\frac{1}{N}\sum^{N}_{j=1}\log{\hat{P}\left[\mathbf{x}^{j}(n)|\mathbf{w}^{j},\Theta\right]}.
\end{equation}
Therefore, we can utilize deep learning approaches to lean the classifier model from the training set $\{\mathbf{x}^{j}(n), \mathbf{w}^{j}\}^{N}_j$.

\subsection{CNN-based SFANC algorithm}
\noindent
Motivated by the work \cite{11}, this paper compares some 1D CNNs and 2D CNNs used for classifying noises in the time domain and frequency domain, respectively. The min-max operation firstly normalizes the input of the network:
\begin{equation}
\hat{x}(n)=\frac{x(n)}{\max [\mathbf{x}(n)]-\min [\mathbf{x}(n)]},
\end{equation}
where $\max [\cdot]$ and $\min [\cdot]$ mean obtaining the maximum and minimum value of $\mathbf{x}(n)$. It aims to rescale the input range into $(-1,1)$ and retain the signal’s negative part that contains phase information. Phase information is quite critical for ANC applications.

\begin{figure}[tp]
\setlength{\abovecaptionskip}{0.cm}
\setlength{\belowcaptionskip}{-0.cm}
\centering
\centerline{\includegraphics[width=0.3\linewidth]{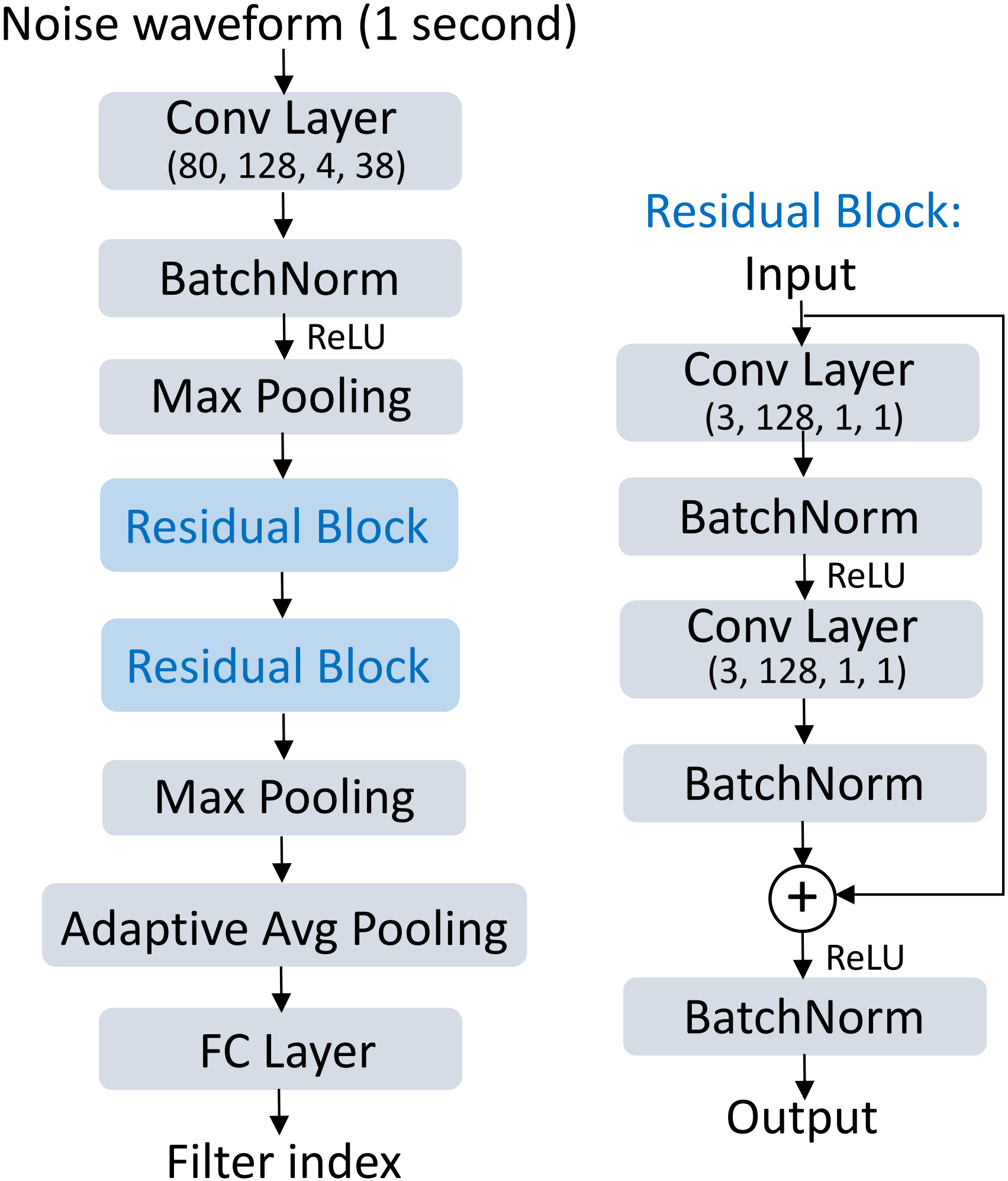}}
\caption{Architecture of the proposed 1D CNN. The configuration of convolution layer is denoted as: (kernel size, channels, stride, padding).}
\label{Fig 3}
\end{figure}

A lightweight 1D CNN illustrated in Figure \ref{Fig 3} is proposed. Every residual block in the network comprises two convolutional layers, subsequent batch normalization, and ReLU non-linearity. Note that a shortcut connection is adopted to add the input with the output in each residual block since residual architecture is demonstrated easy to be optimized \cite{15}. Additionally, the network uses a broad receptive field (RF) in the first convolutional layer and narrow RFs in the rest convolutional layers to fully exploit both global and local information.

\begin{figure}[tp]
\setlength{\abovecaptionskip}{0.cm}
\setlength{\belowcaptionskip}{-0.cm}
\centering
\centerline{\includegraphics[height=3cm,width=0.45\linewidth]{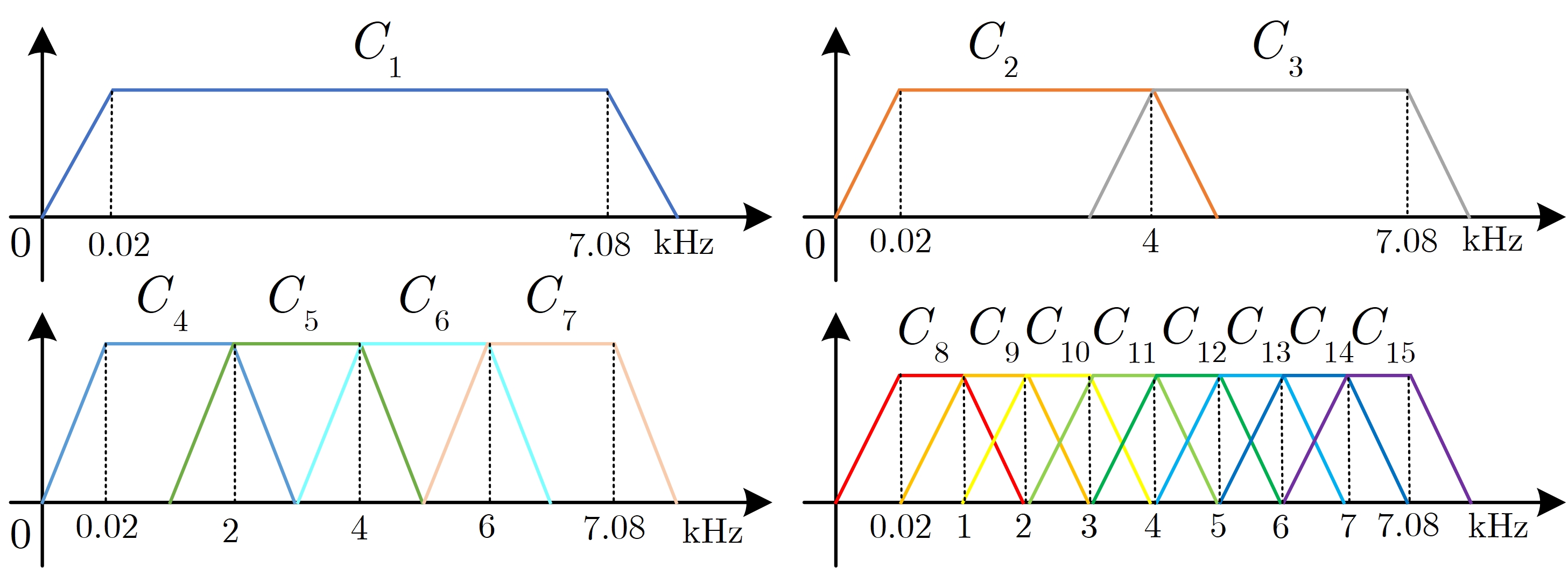}}
\caption{The frequency bands of the noise tracks for pre-training control filters.}
\label{Fig 4}\vspace*{-0.3cm}
\end{figure}

\subsection{Training of CNNs}
\noindent
The primary and secondary paths used in the training stage of the control filters are band-pass filters with a frequency range of $20$Hz-$7,980$Hz. Broadband noises with $15$ frequency ranges shown in Figure \ref{Fig 4} are used to pre-train $15$ control filters. The FxLMS algorithm is adopted to obtain the optimal control filters for these broadband noises due to its low computational complexity. Subsequently, the $15$ pre-trained control filters are saved in the control filter database.

A noise dataset including synthetic and real noise tracks is used in this work. Specifically, $80,000$ synthetic noise tracks and $80,000$ real noise tracks are used for training, $2,000$ real noise tracks for validation, and $2,000$ real noise tracks for testing. The synthetic noise tracks are randomly generated with various frequency bands, amplitudes, and background noise levels.

The SFANC system's sample rate is $16,000$Hz, so each noise track of 1-second duration consists of $16,000$ samples. Each noise track of 1 second duration is taken as primary noise to generate disturbance. The class label of a noise track corresponds to the index of the control filter that achieves the best noise reduction performance on the disturbance.

\section{Experiments}
\noindent
The Adam algorithm was employed to optimize the network during training. The training epoch was set to be $30$. The glorot initialization \cite{16} was used to avoid bursting or vanishing gradients. Additionally, to prevent overfitting, the weights of CNNs were subjected to $\ell_{2}$ regularization with a coefficient of $0.0001$.

\subsection{Comparison of Different Training Schemes}
\noindent
Four different training schemes are compared in training the proposed 1D CNN. The comparison results are summarized in Table \ref{Table 1}. According to Table \ref{Table 1}, training firstly on the synthetic noise tracks and then fine-tuning on the real noise tracks achieves the highest testing accuracy. Noted that simultaneously using synthetic dataset and real dataset for training has not obtained a superior testing accuracy since the characteristics of synthetic noises and real noises are quite different. As discussed above, in the SFANC system, the CNN models can be firstly trained with the synthetic dataset and then fine-tuned with the real noise dataset.

\begin{table}[htp!]
\setlength{\abovecaptionskip}{0.cm}
\setlength{\belowcaptionskip}{-0.cm}
\caption{The performance of different network training schemes.}
\begin{center}
\begin{threeparttable}
\begin{tabular}{|lc|}
\hline
\textbf{Training Scheme} & \textbf{Testing Accuracy} \\
\hline
Only using synthetic dataset & 46.4$\%$ \\
Only using real dataset & 94.6$\%$ \\
Fine-tuning method\tnote{*} & \textbf{95.3$\%$} \\
Simultaneously using synthetic dataset and real dataset & 94.5$\%$ \\
\hline
\end{tabular}
\begin{tablenotes}\scriptsize
\item[*]Training firstly by the synthetic dataset and then fine-tuning by the real dataset.
\end{tablenotes}
\end{threeparttable}
\label{Table 1}
\end{center}\vspace*{-1cm}
\end{table}

\subsection{Comparison of Different Networks}
\noindent
Based on above fine-tuning training scheme, we compared several different 1D networks utilizing raw acoustic waveforms: the proposed 1D CNN, M3 \cite{17}, M5 \cite{17}, M11 \cite{17}, M18 \cite{17}, and M34-res \cite{17}. Also, some light-weight 2D networks including ShuffleNet v2 \cite{18}, MoblieNet v2 \cite{23} and Attention Network \cite{20} are compared in the SFANC method. The performance of these networks on the real testing dataset are summarised in Table \ref{Table 2}.

\begin{table}[htp!]
\setlength{\abovecaptionskip}{0.cm}
\setlength{\belowcaptionskip}{-0.cm}
\caption{Comparisons of different networks used in the SFANC system.}
\begin{center}
\begin{tabular}{|lcc|}
\hline
\textbf{Network} & \textbf{Testing Accuracy} & \textbf{Network Parameters} \\
\hline
\rowcolor{gray!20}
\multicolumn{3}{|c|}{1D Convolutional Neural Networks} \\
\hline
Proposed 1D Network & \textbf{95.3$\%$} & \textbf{0.21M} \\
M3 Network & 93.7$\%$ & 0.22M \\
M5 Network & 94.9$\%$ & 0.56M \\
M11 Network & 94.5$\%$ & 1.79M \\
M18 Network & 93.8$\%$ & 3.69M \\
M34-res Network & 94.4$\%$ & 3.99M \\
\hline
\rowcolor{gray!20}
\multicolumn{3}{|c|}{2D Convolutional Neural Networks} \\
\hline
ShuffleNet v2 & \textbf{95.5$\%$} & \textbf{0.25M} \\
MoblieNet v2 & 95.6$\%$ & 2.89M \\
Attention Network & 94.9$\%$ & 4.95M \\
\hline
\end{tabular}
\label{Table 2}
\end{center}\vspace*{-0.5cm}
\end{table}

As shown in Table \ref{Table 2}, the proposed 1D network obtains the highest classification accuracy of $95.3\%$ with the fewest network parameters among the 1D networks. As for 2D networks, the ShuffleNet v2 achieves a similar classification accuracy as MoblieNet v2 and requires far fewer parameters. By considering both the testing accuracy and the number of parameters, the ShuffleNet v2 performs best on the testing dataset among the 2D networks. Compared to the proposed 1D network, the ShuffleNet v2 obtains a slight improvement in classification accuracy but requires a little more network parameters. Therefore, it is found that the proposed 1D network and ShuffleNet v2 perform better in classifying noises in the SFANC system. The two light-weight networks can be implemented on mobile platforms, but using acoustic models directly from the raw waveform data is more convenient \cite{24}. Hence, the proposed 1D network is preferred.

\subsection{Non-stationary Noise Cancellation}
\begin{figure}[htp!]
\setlength{\abovecaptionskip}{0.cm}
\setlength{\belowcaptionskip}{-0.cm}
\centering
\subfigure{
\includegraphics[width=0.4\linewidth]{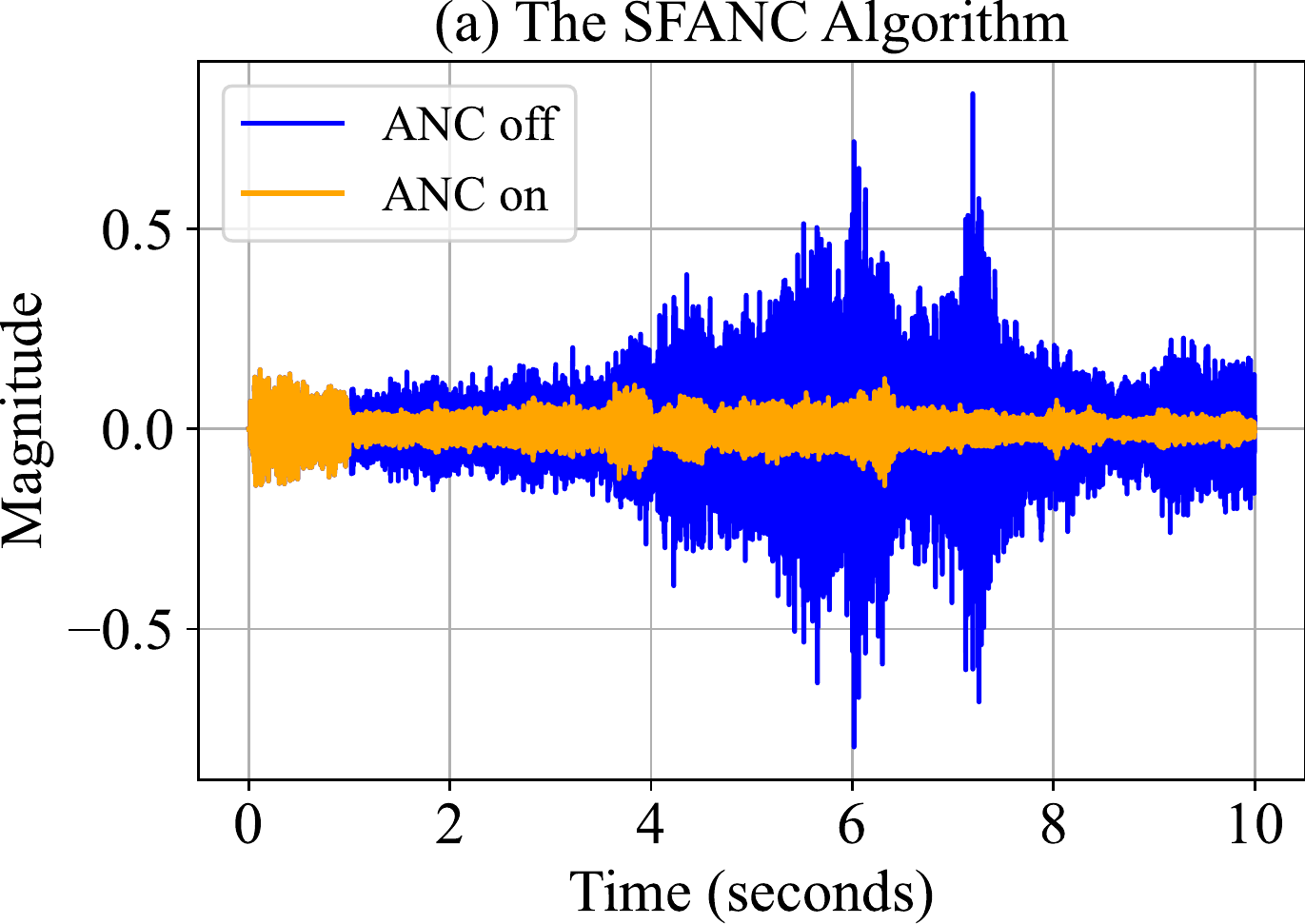}
}
\subfigure{
\includegraphics[width=0.4\linewidth]{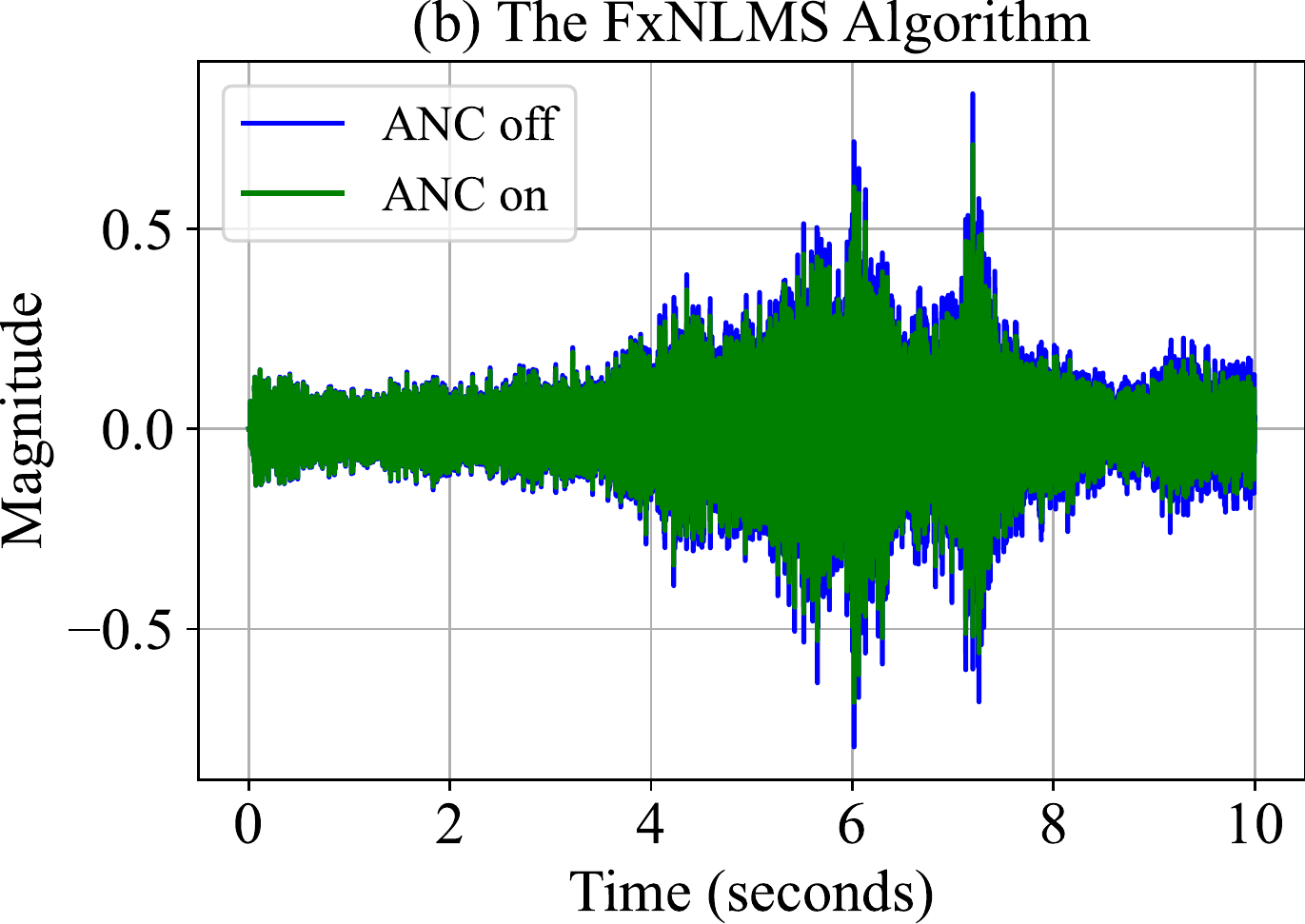}
}
\subfigure{
\includegraphics[width=0.4\linewidth]{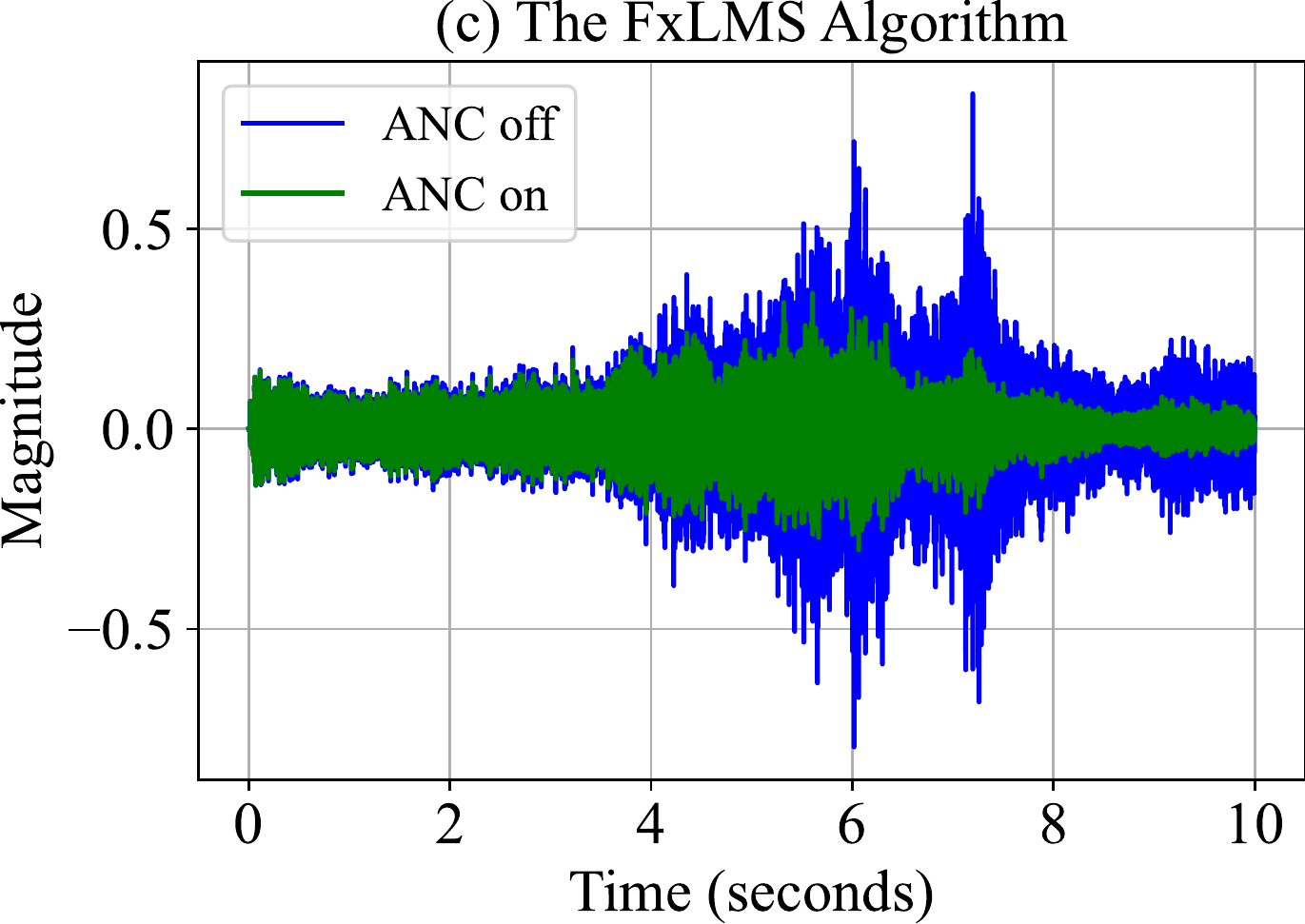}
}
\subfigure{
\includegraphics[width=0.4\linewidth]{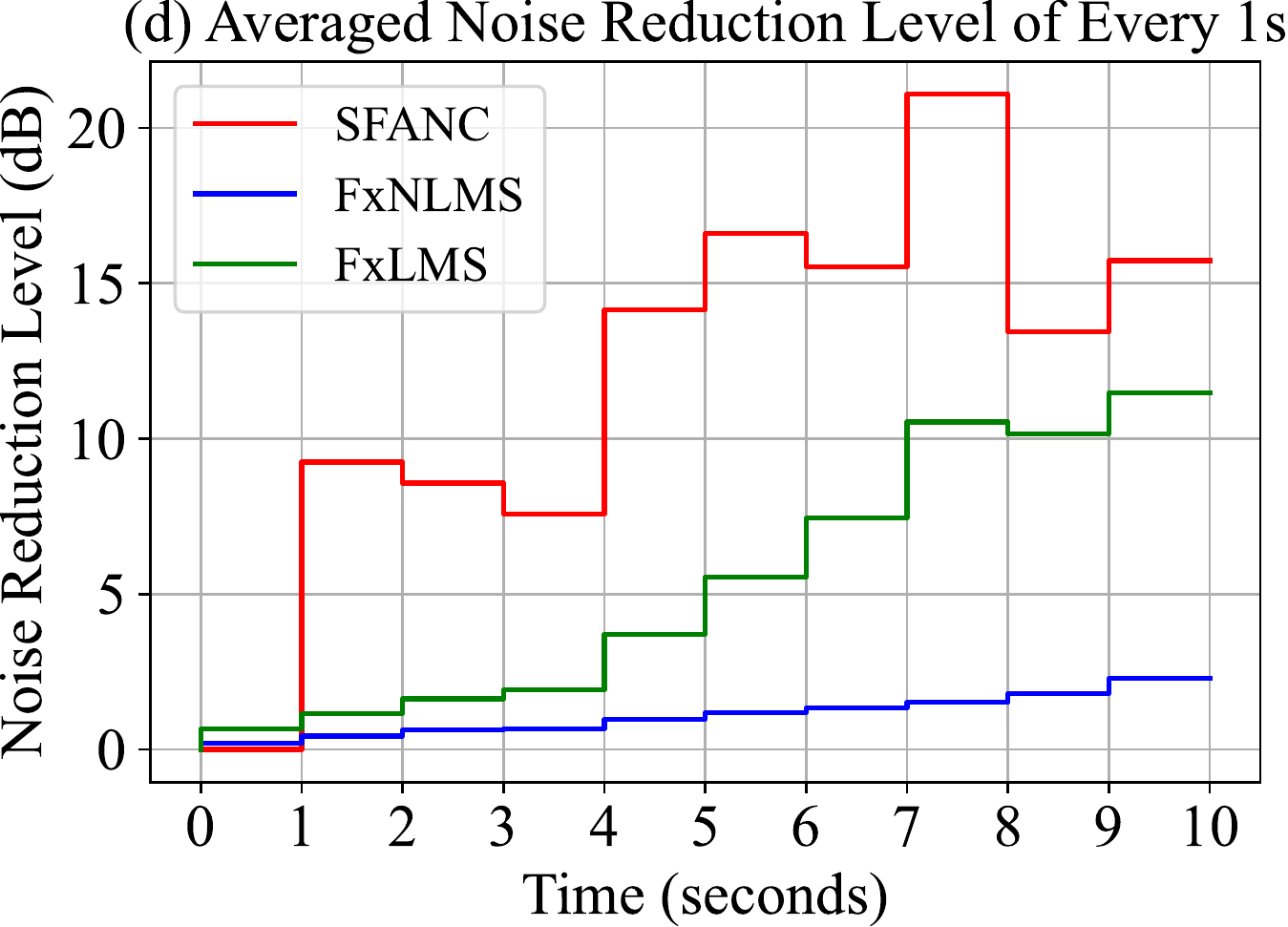}
}
\caption{(a)-(c): Error signals of different ANC algorithms, (d): Averaged noise reduction level of every 1 second, on the aircraft noise.}
\label{Fig 5}
\end{figure}

\noindent
This section uses the SFANC algorithm based on the proposed 1D network, FxLMS algorithm, and FxNLMS algorithm to attenuate a recorded aircraft noise. The aircraft noise is non-stationary and has a frequency range of 50Hz-14,000Hz. It does not belong to the training dataset. The step size of the FxLMS and FxNLMS algorithm is set to $0.0001$, and the control filter length is $1,024$ taps. The noise reduction results using different ANC methods on the aircraft noise are shown in Figure \ref{Fig 5}.

From the results in Figure \ref{Fig 5}, we can observe that the SFANC method responds to the aircraft noise much faster than the FxLMS and FxNLMS algorithms. Also, the SFANC method consistently outperforms the FxLMS and FxNLMS algorithm in the noise reduction process. In particular, during $1$s-$2$s, the averaged noise reduction level achieved by the SFANC algorithm is about 7dB and 8dB more than that of FxLMS and FxNLMS, respectively. Therefore, the results on the aircraft noise confirm that the SFANC method can rapidly select the most suitable pre-trained control filter given the noise type. In contrast, adaptive algorithms show slow responses to the aircraft noise due to adaptive updating.

\section{CONCLUSIONS}
\noindent
Active noise control (ANC) technologies have been widely used to deal with low-frequency noises. However, adaptive ANC algorithms are typically limited by slow convergence speed. In this paper, CNNs are used to automatically select the best pre-trained control filters given different noises. Also, light-weight CNNs implemented on a co-processor can decouple the computational load from the real-time noise controller. Numerical simulations show that the CNN-based SFANC method improves response time while maintaining low computational complexity and high robustness. Additionally, the effectiveness of the proposed 1D network and the fine-tuning training strategy are confirmed in the SFANC method. In future works, we will explore more efficient and robust ANC algorithms based on deep learning.

\bibliographystyle{unsrt}
\bibliography{sample}

\end{document}